%% file: main.tex
\documentclass[runningheads]{llncs}
\usepackage{graphicx}

\usepackage{tikz}
\usepackage{comment}
\usepackage{amsmath,amssymb} 
\usepackage{color}
\usepackage[normalem]{ulem}

\usepackage[accsupp]{axessibility}  

\definecolor{citecolor}{HTML}{0071bc}
\usepackage[pagebackref=true,breaklinks=true,letterpaper=true,colorlinks,citecolor=citecolor,bookmarks=false]{hyperref}
\usepackage{xspace}
\usepackage[capitalize]{cleveref}
\crefname{section}{Sec.}{Secs.}
\Crefname{section}{Section}{Sections}
\Crefname{table}{Table}{Tables}
\crefname{table}{Tab.}{Tabs.}

\DeclareMathOperator*{\argmax}{arg\,max}
\newcommand{\etal}{\textit{et al}.}
\newcommand{\ie}{i.e. }
\newcommand{\mvd}{\textsc{MvDeCor}\xspace}

\begin{document}
\pagestyle{headings}
\mainmatter
\def\ECCVSubNumber{525}  

\title{\textsc{MvDeCor}: Multi-view Dense Correspondence Learning for Fine-grained 3D Segmentation}

\titlerunning{\textsc{MvDeCor}}
%
\author{Gopal Sharma\inst{1}\thanks{The work was mostly done during Gopal's internship at NVIDIA} \and
Kangxue Yin\inst{2} \and
Subhransu Maji\inst{1} \and
Evangelos Kalogerakis\inst{1} \and
Or Litany\inst{2} \and
Sanja Fidler\inst{2,3,4}
}
\authorrunning{G. Sharma et al.}
%
\institute{University of Massachusetts, Amherst \and NVIDIA \and  University of Toronto \and Vector Institute}
\maketitle
\begin{abstract}
\input{abstract}
\end{abstract}
\input{introduction}

\input{related}

\input{method}
 \input{experiments}

\input{conclusion}

\newpage

\bibliographystyle{splncs04}
\bibliography{egbib}

\input{supp}

\end{document}

%% file: abstract.tex
We propose to utilize self-supervised techniques in the 2D domain for fine-grained 3D shape segmentation tasks. 
This is inspired by the observation that view-based surface representations are more effective at modeling high-resolution surface details and texture than their 3D counterparts based on point clouds or voxel occupancy.
Specifically, given a 3D shape, we render it from multiple views, and set up a dense correspondence learning task within the contrastive learning framework. 
As a result, the learned 2D representations are view-invariant and geometrically consistent, leading to better generalization when trained on a limited number of labeled shapes compared to alternatives that utilize self-supervision in 2D or 3D alone.
Experiments on textured (RenderPeople) and untextured (PartNet) 3D datasets show that our method outperforms state-of-the-art alternatives in fine-grained part segmentation. 
The improvements over baselines are greater when only a sparse set of views is available for training or when shapes are textured, indicating that \mvd benefits from both 2D processing and 3D geometric reasoning. Project page: \href{https://nv-tlabs.github.io/MvDeCor/}{https://nv-tlabs.github.io/MvDeCor/}

%% file: introduction.tex
\section{Introduction}
Part-level interpretation of 3D shapes is critical for many applications in computer graphics and vision, including 3D content editing, animation,  simulation and synthesizing virtual datasets for visual perception, just to name a few.
Specifically, our goal in this work is to perform fine-grained shape segmentation from limited available data. This poses two main challenges. First, training deep networks relies on large-scale labeled datasets that require tremendous annotation effort. For this reason, previous methods have proposed self-supervised feature extraction, however these mostly rely on point cloud or voxel-based networks. This brings us to the second challenge -- these 3D networks have a limited ability to capture fine-grained surface details in their input points or voxels due to the limits on the sampling density.

\begin{figure}[t]
\centering
\includegraphics[width=0.9\linewidth]{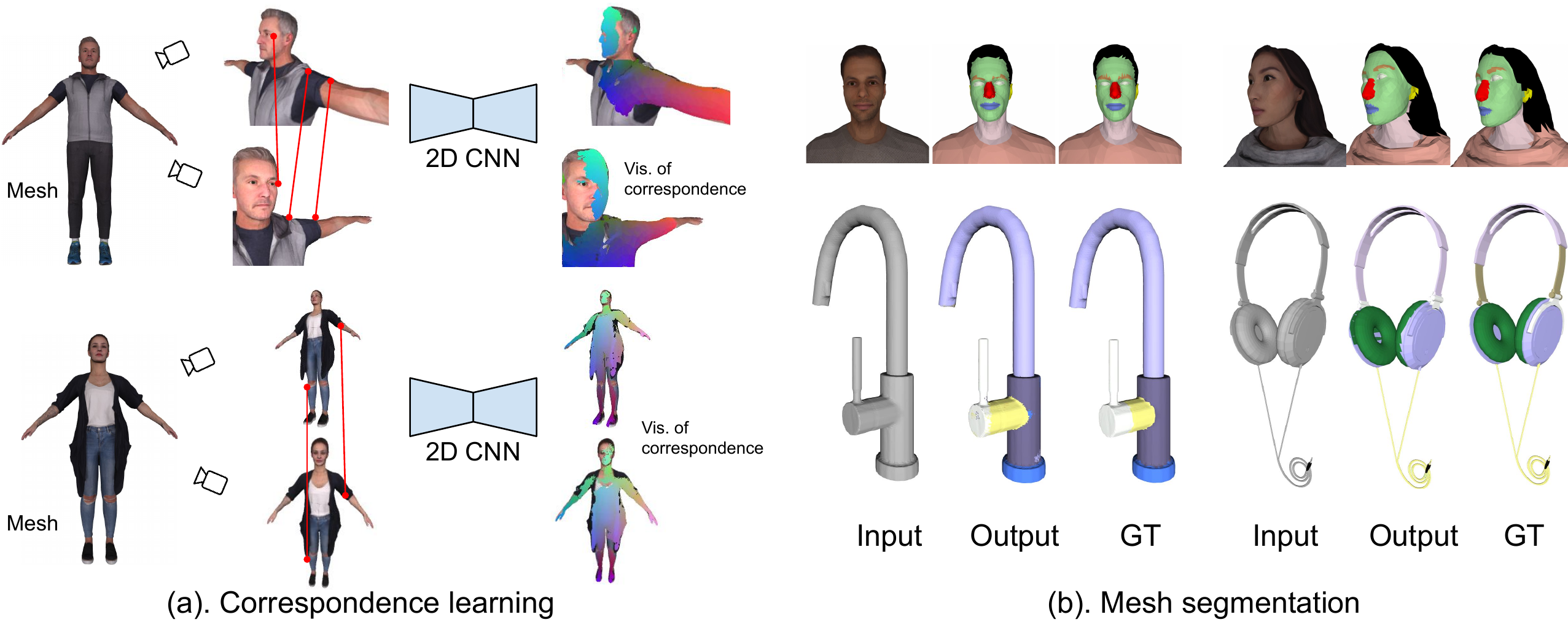}
\caption{
\textbf{The \mvd pipeline}. 
(a) Dense 2D representations are learned using pixel-level correspondences guided by 3D shapes. 
(b) The 2D representations can be fine-tuned using a few labels for 3D shape segmentation tasks in a multi-view setting.}
\label{fig:teaser}
\end{figure}

We present $\textsc{MvDeCor}$, a self-supervised technique for learning dense 3D shape representations based on the task of learning correspondences across views of a 3D shape (Fig.~\ref{fig:teaser}).
At training time we render 3D shapes from multiple views with known correspondences and setup a contrastive learning task to train 2D CNNs. In doing so, we take advantage of the excellent abilities of 2D networks to capture fine details. 
The learned 2D representations can be directly used for part segmentation on images, or projected onto the shape surface to produce a 3D representation for 3D tasks (Fig.~\ref{fig:teaser}~\&~\ref{fig:overview}).
The approach works well in standard few-shot fine-grained 3D part segmentation benchmarks, outperforming prior work based on 2D and 3D self-supervised learning (\S\ref{sec:method}, Tab.~\ref{table:partnetfewshape}~\&~\ref{table:partnetfewviews}).

Many previous representation learning methods for 3D shapes are based on self-reconstruction loss~\cite{latentpc,yang2018foldingnet,atlasnet,mrt18,chen2019bae} or contrastive learning ~\cite{PointContrast2020,zhang_depth_contrast,hou2021exploring}
where point clouds and voxels are the main choices of 3D data formats.
In contrast, our work is motivated from the observation that view-based surface representations are more effective at modeling high-resolution surface details and texture than their 3D counterparts based on point clouds or voxel occupancy. We also benefit from recent advances in network architectures and self-supervised learning for 2D CNNs. 
In addition, our approach allows training the network using 2D labeled views rather than fully labeled 3D shapes. This is particularly beneficial because annotating 3D shapes for fine-grained semantic segmentation is often done using 2D projections of the shape to avoid laborious 3D manipulation operations \cite{Yi:2016:SAF}.

Compared to techniques based on 3D self-supervision, \mvd demonstrates significant advantages.
On the PartNet dataset~\cite{partnet} with fine-grained (Level-3) parts, our method achieves 32.6\% mIOU compared to a PointContrast~\cite{PointContrast2020}, a self-supervised learning technique that achieves 31.0\% mIOU (Tab.~\ref{table:partnetfewshape}).
While some of the benefit comes from the advantages of view-based representations, e.g., off-the-self 2D CNNs trained from scratch outperform their 3D counterparts, this alone does not explain the performance gains. 
\mvd outperforms both ImageNet pretrained models (29.3\% mIOU) and dense contrastive learning~\cite{wang2020DenseCL} (30.8\% mIOU), suggesting that our correspondence-driven self-supervision is beneficial. 
These improvements over baselines are even larger when sparse view supervision is provided — \mvd generalizes to novel views as it has learned a view invariant local representations of shapes (Tab. \ref{table:partnetfewviews}).

We also present experiments on the RenderPeople~\cite{Renderpeople} dataset consisting of textured 3D shapes of humans, which we label with 13 parts (\S\ref{sec:exp}). We observe that 2D self-supervised techniques performs better than their 3D counterparts, while \mvd offers larger gains over both 2D and 3D baselines (Tab.~\ref{table:renderpeople}). Surprisingly on this dataset we find that when texture is available, the view-based representations do not require the use of depth and normal information, and in fact the models generalize better without those, as explained in \S\ref{sec:renderpeople}. \mvd gives $17.3\%$ mIOU improvement over training a network from scratch when only a few labeled examples are provided for supervision.

To summarize, we show that multi-view dense correspondence learning induces view-invariant local representations that generalize well on few-shot 3D part segmentation tasks. 
Our approach \mvd outperforms state-of-the art 2D contrastive learning methods, as well as 3D contrastive learning methods that operate on point cloud representations. 
After a discussion of prior work on 2D and 3D self-supervised learning in \S\ref{sec:relatedworks}, we describe our method and experiments in \S\ref{sec:method} and \S\ref{sec:exp} respectively.

%% file: related.tex
\section{Related Works}\label{sec:relatedworks}
Our work lies at the intersection of 3D self-supervision, 2D self-supervision, and multi-view representations. 

\paragraph{\textbf{3D self-supervision.}}
Many self-supervised approaches in 3D shape are based on training an autoencoder with a reconstruction loss.
For example, Achlioptas \etal~\cite{latentpc} train a PointNet~\cite{qi2017pointnet} with a Chamfer or EMD loss.
FoldingNet~\cite{yang2018foldingnet} deforms a 2D grid using a deep network conditioned on the shape encoding to match the output shape. AtlasNet~\cite{atlasnet} uses multiple decoders to reconstruct the surface as a collection of patches. 
BAE-NET~\cite{chen2019bae} splits reconstruction across decoding branches, but adopted an implicit field shape representation instead of point clouds. 
Once trained the representations from the encoder can be used for downstream tasks.
%
Alternatives to reconstruction include prediction based on $k$-means~\cite{hassani2019unsupervised}, convex decomposition~\cite{selfsupacd,acd}, primitive fitting \cite{prifit2022,stardomain2020,abstractionTulsiani17} and 3D jigsaw puzzles~\cite{sauder2019self,jointssl}.
Unsupervised learning for recovering dense correspondences between non-rigid shapes has been studied in \cite{halimi2018selfsupervised,ginzburg2019cyclic}, however it relies on a near-isometry assumption that does not fit clothed people and furniture parts, used in our work.
We instead use partial correspondences from the 3D models to supervise 2D networks. Wang \etal~\cite{wang2020few} proposed a deep deformation approach that aligns a labeled template shape to unlabeled target shapes to transfer labels. However this method is not effective for
fine-grained segmentation of shapes as shown in \S\ref{sec:exp}, since deformation often distorts surface details. A few recent works \cite{PointContrast2020,wang2021unsupervised,zhang_depth_contrast} have learned
per-point representations for point clouds under a contrastive learning framework.
Networks pre-trained in this way are further fine-tuned for 3D downstream tasks.
However, point cloud based shape representations limit the ability to capture fine-grained details and texture.

\begin{figure*}[t]
\centering
\includegraphics[width=0.7\textwidth]{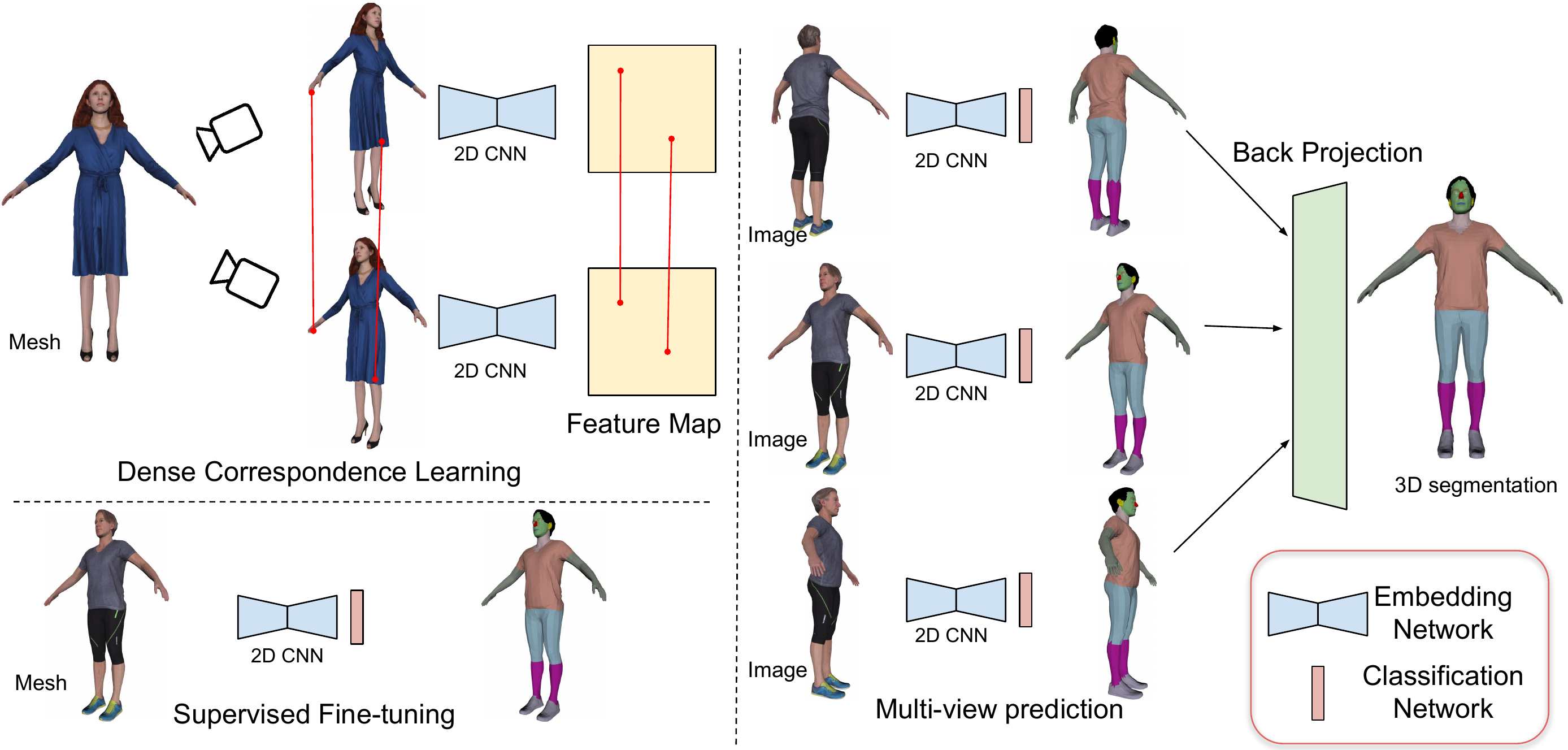}
\caption{\textbf{Overview of \mvd.} \emph{Top left:} Our self-supervision approach takes two overlapping views (RGB image, with optional normal and depth maps) of a 3D shape and passes it through a network that produces per-pixel embeddings. We define a dense contrastive loss promoting similarity between matched pixels and minimizing similarity between un-matched pixels. \emph{Bottom left:} Once the network is trained we add a segmentation head and fine-tune the entire architecture on a few labeled examples to predict per-pixel semantic labels. \emph{Right:} Labels predicted by the 2D network for each view are back-projected to the 3D surface and aggregated using a voting scheme.
}\label{fig:overview}
\end{figure*}
\paragraph{\textbf{2D self-supervision.}}
While early work focused on training networks based on proxy tasks such as image colorization, rotation prediction, and jigsaw puzzles, contrastive learning \cite{wu2018unsupervised,he2019moco,chen2020simple,NEURIPS2020_f3ada80d} has emerged as a popular technique.
Most of these representations are based on variants of InfoNCE loss~\cite{oord2018representation}, where the mutual information between two views of an image obtained by applying synthetic transformations is maximized.
DenseCL~\cite{wang2020DenseCL} modifies the contrastive approach to include information across locations within an image to learn dense representations. We use this method as the representative 2D self-supervised baseline.
However, the above methods  work on the 2D domain and lack any 3D priors incorporated either in the network or
in the training losses. 
Correspondence learning has been used as self-supervision task to learn local descriptors for geometric matching in structure from motion applications~\cite{sarlin20superglue,wang2020DenseCL,hou2021pri3d,LuoStereo}.
However, much of this work has focused on instance matching, while our goal is to generalize across part instances within a category.
The most related work to ours is Pri3D~\cite{hou2021pri3d} that also proposes to learns geometry-aware embedding with a contrastive loss based on pixel correspondences across views. 
Their work focuses on improving 2D representations using 3D supervision for 2D tasks such as scene segmentation and object detection, while we deal with fine-grained 3D segmentation.

\paragraph{\textbf{Multi-view representation.}}
Our method is motivated by earlier multi-view approaches for 3D shape recognition and segmentation~\cite{Huang:2017:LMVCNN,kalogerakis2017shapepfcn,su15mvcnn,choy_nips16,kundu2020virtual}. In these approaches, 
multiple views of the shapes are processed by a 2D network to obtain pixel-level representations. These are back-projected to the 3D shape to produce 3D surface representations. More recently, Kundu \etal \cite{kundu2020virtual} applies multi-view
fusion for 3D semantic segmentation of scenes.
Genova et al. \cite{genova2021learning} leverage a pretrained 2D segmentation network to synthesize pseudo-labels for training a 3D segmentation network.
The above approaches benefit from large-scale pretraining on image dataset,
and the ability of 2D CNNs to handle higher image resolutions compared to voxel grids and point clouds. They continue to outperform 3D deep networks on many 3D tasks (e.g.,~\cite{su2018deeper,goyal2021revisiting}). Semantic-NeRF \cite{zhi2021place} jointly reconstructs geometry and semantics of a 3D scene from multi-view images with partial or noisy labels.
All the above view-based methods are trained in a supervised or weakly supervised manner, while ours is based on self-supervision.

%% file: method.tex
\section{Method}\label{sec:method}
Our goal is to learn a multi-view representation for 3D shapes for the task of fine-grained 3D shape segmentation.
For this task, we assume a large dataset of unlabeled 3D shapes and a small number of labeled examples. For the latter, we consider two settings (1) when labels are provided on the surface of the 3D shape, and (2) labels are provided on the images (projections) of the 3D shape.
To that end, we use multi-view dense correspondence learning for pre-training, followed by a feature fine-tuning stage on the downstream segmentation task. 
In the pre-training stage described in \S \ref{sec:mvpretrain}, we have a set of unlabeled 3D shapes (either with or w/o textures) 
from which we render 2D views, and build ground-truth dense correspondences between them. After pre-training on the dense correspondence learning task, the network learns pixel-level features that are robust to view changes and is aware of fine-grained details.

In the fine-tuning stage (\S \ref{sec:segmentation}), we train a simple convolutional head on top of the pixel-level embeddings, supervised by a small number of annotated examples, to segment the multi-view renderings of the 3D shapes. The network pre-trained in this fashion produces better segmentation results under the few-shot semantic segmentation regime in comparison to baselines. 
We aggregate multi-view segmentation maps onto 3D surface via an entropy-based voting scheme (\S \ref{sec:technicaldetails}). Figure~\ref{fig:overview} shows the overview of our approach.

\subsection{Multi-view dense correspondence learning\label{sec:mvpretrain}}

Let us denote the set of \textit{unlabeled} shapes as $\mathcal{X}_u$. Each shape instance \mbox{$X \in \mathcal{X}_u$} can be rendered from a viewpoint $i$ into color, normal and depth images denoted as $V^i$. 
We use a 2D CNN backbone $\mathbf{\Phi}$ which maps each view into pixel-wise embeddings $\{\mathbf{\Phi}(V^i)_{p}\} \in \mathbb{R}^D$, where $p$ is an index of a pixel and $D$ is the dimensionality of the embedding space. 
We pre-train the network  ${\mathbf{\Phi}}$ using the following self-supervised loss:
\begin{equation}\label{eq:ssl}
    {\cal L}_{\textsf{ssl}}  =  \mathop{\mathbb{E}}_{ \substack{ V^i,V^j \sim \mathcal{R}({X}) \\ X \sim \mathcal{X}_u } } \Big[\ell_{\textsf{ssl}}\big(\mathbf{\Phi}(V^i), \mathbf{\Phi}(V^j) \big) \Big]
\end{equation}
where $\mathcal{R}({X})$ is the set of 2D renderings of shape $X$,  and $V^i$ and $V^j$ are sampled renderings in different views from  $\mathcal{R}({X})$.
The self-supervision loss $\ell_{\textsf{ssl}}$ is applied to the pair of sampled views $V^i$ and $V^j$.

Since $V^i$ and $V^j$ originate from the 3D mesh, each foreground pixel in the rendered images corresponds to a 3D point on the surface of a 3D object.  We find matching pixels from $V^i$ and $V^j$ when their corresponding points in 3D lie within a small threshold radius.
We use the obtained dense correspondences in the self-supervision task. Specifically, we train the network to minimize the distance between pixel embeddings that correspond to the same points in 3D space and maximize the distance between unmatched pixel embeddings. This encourages the network to learn pixel embeddings to be invariant to views, which is a non-trivial task, as two rendered views of the same shape may look quite different, consisting of different contexts and scales.
We use InfoNCE~\cite{oord2018representation} as the self-supervision loss. 
Given two rendered images ($V^i$ and $V^j$) from the same shape $X$ and pairs of matching pixels $p$ and $q$, the InfoNCE loss is defined as:
\begin{equation}\label{eq:infonce}
    \ell_{\textsf{ssl}}\big( \mathbf{\Phi}(V^i), \mathbf{\Phi}(V^j)  \big) =  
     -\sum_{(p,q) \in M} \log\frac{\exp \big(\mathbf{\Phi}(V^{i})_p \cdot \mathbf{\Phi}(V^{j})_q / \tau\big)}{\sum_{(., k)\in M }\exp\big(\mathbf{\Phi}(V^i)_p \cdot \mathbf{\Phi}(V^j)_k / \tau\big)},
\end{equation}
where $\mathbf{\Phi}(V^i)_p$ is the embedding of pixel $p$ in view $i$, 
$M$ is the set of paired pixels between two views that correspond to the same points in 3D space, and the temperature is set to $\tau=0.07$ in our experiments.  We use two views that have at least $15\%$ overlap.
The output $\mathbf{\Phi}(V^i)_p$ and $\mathbf{\Phi}(V^j)_q$ of the embedding module are normalized to a unit hyper-sphere. Pairs of matching pixels are treated as positive pairs. The
above loss also requires sampling of negative pairs. Given the matching pixels $(p, q) \in M$ from $V^{i}$ and $V^{j}$ respectively, for each 
pixel $p$ from the first view, the rest of the pixels $k\neq q $  appearing in $M$ and belonging to the second view, yield the negative pixel pairs $(p, k)$.

\subsection{Semantic segmentation of 3D shapes}\label{sec:segmentation}
In the fine-grained shape segmentation stage, the network learns to predict pixel level segmentation labels.
Once the embedding module is pre-trained using the self-supervised approach, it is further fine-tuned in the segmentation stage, using a small labeled shape set $\mathcal{X}_l$ to compute a supervised loss, as follows:
\begin{align}\label{eq:joint}
 \min_{\mathbf{\Phi}, \mathbf{\Theta}} \lambda {\cal L}_{\textsf{ssl}} + {\cal L}_{\textsf{sl}} \text{, where } 
 {\cal L}_{\textsf{sl}} = \mathop{\mathbb{E}}_{(X, Y) \sim \mathcal{X}_l}  \Big[
    \mathop{\mathbb{E}}_{(V^i,L^i) \in \mathcal{R}(X, Y)}\ell_{\textsf{sl}}\big(L^i,\mathbf{\Theta} \circ \mathbf{\Phi} (V^i)\big) \Big], 
\end{align}
and $\mathbf{\Theta}$ is the segmentation module, $\lambda$ is a hyper-parameter set to $0.001$, and $\ell_{\textsf{sl}}$ is the semantic segmentation loss implemented using cross-entropy loss applied to each view of the shape separately.  $\mathcal{R}(X, Y)$ is the set of renderings for shape $X$ and its 3D label map $Y$.  $L^i$ represents the projected labels from the 3D shape for view $V^i$.
Since the labeled set is much smaller than the unlabeled set, the network could overfit to the small set. To avoid this over-fitting during the fine-tuning stage, we use the self-supervision loss ${\cal L}_{\textsf{ssl}}$ as an auxiliary loss along with supervision loss ${\cal L}_{\textsf{sl}}$ as is shown in Eq. \ref{eq:joint}. In Table \ref{table:viewselection} we show that incorporating this regularization improves the performance.

During inference we render multiple overlapping views of the 3D shape and segment each view. The per-pixel labels are then projected onto the surface. We use ray-tracing to encode the triangle index of the mesh for each pixel. To aggregate labels from different views for each triangle, one option is to use majority-voting. An illustration of the process is shown in Fig.~\ref{fig:overview}. However, not all views should contribute equally towards the final label for each triangle as some views may not be suitable to recognize a particular part of the shape. We instead define a weighted voting scheme based on the average entropy of the probability distribution predicted by the network for a view. Specifically, a weight \mbox{$W^{(i)} = 
(1 - \sum_{p \in F^{(i)}} H^{(i,p)} / |F^{(i)}|) ^ \gamma$} is given to the view $i$, where $F^{(i)}$ is its set of foreground pixels, $H^{(i,p)}$ is the entropy of the probability distribution predicted by the network at pixel $p$, and $\gamma$ is a hyperparameter set to $20$ in our experiments. More weight is given to the view with less entropy. Consequently, for each triangle $t$ on the mesh of the 3D shape, the label is predicted as: \mbox{$l_t = \argmax_{c \in C} \sum_{i \in I, p \in t} W^{(i)} P^{(i,p)}$}, 
where $I$ is the set of views where the triangle $t$ is visible, $P^{(i,p)}$ is the probability distribution of classes at a pixel $p \in t$ in view $i$, and $C$ is the set of segmentation classes. 
In cases where no labels are projected to a triangle due to occlusion, we assign a label to the triangle by nearest neighbor search.

\subsection{Implementation details}\label{sec:technicaldetails}

The embedding $\mathbf{\Phi}$ is implemented as the DeepLabV3+ network~\cite{deeplabv3plus2018} originally proposed for image segmentation with ResNet-50 backbone. We add extra channels in the first layer to incorporate depth and normal maps. Specifically, it takes a K-channel image ($V^i$) as input of size $H\times W\times K$ and outputs $\mathbf{\Phi}(V^i)$  per pixel features of size $H\times W\times 64$, where the size of pixel embedding is $64$.  
In the second stage, we add a segmentation head (a 2D convolutional layer with a softmax) on top of the pixel embedding network to produce per-pixel semantic labels. Additional architecture details are provided in the supplementary material.

To generate the dataset for the self-supervision stage, we start by placing a virtual camera at 2 unit radius around the origin-centered and unit normalized mesh. We then render a fixed number of images by placing the camera at uniform positions and adding random small perturbations in the viewing angle and scale. In practice, we use approximately $90$ rendered images per shape to cover most of the surface area of the shapes. We also render depth  and normal maps for each view. Normal maps are represented in a global coordinate system. Depth maps are normalized within each view. We use ray tracing to record the triangle index to which each pixel corresponds to and also the point-of-hit for each pixel. This helps in identifying correspondences between two views of the same shape. More information is provided in the supplementary material. 

%% file: experiments.tex
\section{Experiments and results}\label{sec:exp}
\subsection{Dataset}
\noindent
We use the following datasets in our experiments, samples from which are visualized in the Supplement. The license information is provided in the Supplement.
\paragraph{\textbf{PartNet}~\cite{partnet}.} This dataset provides \textit{fine-grained} semantic
  segmentation annotation for various 3D shape categories, unlike
  the more coarse-level shape parts in the ShapeNet-Part dataset.  We
  use 17 categories from ``level-3'', which denotes the finest
  level of segmentation. On average the categories contain $16$ parts, ranging from $4$ for the Display category, to $51$ for Table category. For training in the few-shot framework, we use the entire training and validation set as
  the self-supervision dataset $\mathcal{X}_u$, and select $k$ shapes from the train set as labeled dataset
  $\mathcal{X}_l$ for the fine-tuning stage.

\paragraph{\textbf{RenderPeople}~\cite{Renderpeople}.} This dataset contains 1000 human
textured models represented as triangle meshes in two poses. We use
936 shapes of them for self-supervision. We label the remaining 64 shapes
with 13 different labels, while focusing more on facial semantic parts. 
The 64 labeled shapes consist of 32 different identities in 2 poses.
We randomly split the 32 identities into 16 and 16, so that we get 32 shapes as the labeled training set, and 32 shapes as the test set for evaluation. More details of the labeling and individual semantic parts are provided in the Supplement.

\paragraph{\textbf{ShapeNet-Part}~\cite{Yi:2016:SAF}.} We also show experiments on ShapeNet-Part dataset  in the Supplementary material where we outperform previous works according to the class-average mIOU metric.
\subsection{Experiment settings}\label{sec:expsetting}
\paragraph{\textbf{Segmentation using limited labeled shapes.}}
We pre-train our 2D embedding network and fine-tune it with a segmentation head using $k$ labeled shapes. 
During fine-tuning, each shape is rendered from 86 different views.  We render extra 10 images for the RenderPeople dataset that focuses more on details of facial regions. Each view consists of a grayscale image for PartNet dataset and textured image for RenderPeople dataset, a normal and a depth map for both datasets. For PartNet dataset, pre-training is done using all shape categories and fine-tuning is done on individual category specific manner.
All experiments in this few-shot setting are run $5$ times on randomly selected $k$ labeled training shapes and the average part mIOU is reported. 

\paragraph{\textbf{Segmentation using limited labeled views per shape.}}
In this setting supervision is available in 2D domain in the form of sparse set of labeled views per shape. 
Specifically, a small number of $k$ shapes are provided with a small number of $v$ labeled views per shape.
For training 3D baselines, labeled views are projected to 3D mesh and corresponding points are used for supervision.
Similar to the first setting, all experiments are run $5$ times on randomly selected $k$ labeled training shapes and the average part mIOU is reported. 

\subsection{Baselines}\label{sec:baselines}
We compare our method against the following baselines:
\begin{itemize}
\item \textbf{(2D) Scratch.} In this baseline, we train our 2D networks ($\mathbf{\Phi}$ and segmentation head $\mathbf{\Theta}$) directly using  the small set of labeled examples, without pre-training, and apply the same multi-view aggregation.

\item  \textbf{(2D) ImageNet.} We create a baseline in which the backbone ResNet in our
embedding network $\mathbf{\Phi}$  is initialized with ImageNet pre-trained weights and the
entire network is fine-tuned using few-labeled examples. 
The first layer of ResNet trained on Imagenet is adapted to take the extra channels
(normal and depth maps) following the method proposed in ShapePFCN
\cite{kalogerakis2017shapepfcn}.   

\item \textbf{(2D) DenseCL.} To compare with the 2D dense contrastive
learning method we also create a baseline using DenseCL
\cite{wang2020DenseCL}. We pre-train this method on our unlabeled
dataset $\mathcal{X}_u$ using their original codebase. Once this network is trained,
we initialize the backbone network with the pre-trained weights and
fine-tune the entire architecture using the available labeled set. Further
details about training these baselines are in the
Supplement.

\item \textbf{(3D) Scratch.}  In this baseline, we train a 3D ResNet based on sparse convolutions (Minkowski
Engine \cite{choy20194d}) that takes uniformly sampled points and their normals from the surface and
predicts semantic labels for each point. We train this network directly using the small set of labeled examples.

\item \textbf{(3D) PointContrast.} 
To compare with the 3D self-supervision methods, we
pre-train the above 3D ResNet (Minkowski Engine) on $\mathcal{X}_u$ using the approach proposed in PointContrast
\cite{PointContrast2020}. The pre-trained network is later
fine-tuned by adding a segmentation head (a 3D convolution layer and softmax) on top to predict per-point
semantic labels. We use the codebase provided by authors to train the network.  Details are provided in the  Supplement.
\item \textbf{(3D) Weak supervision via learned 
deformation.} We use the approach  by Wang \etal~\cite{wang2020few}, which uses 
learning-based deformation and transfer of labels from the
labeled set to unlabeled shapes. We use our
labeled and unlabeled set to train this method using the code provided by the authors.
\end{itemize}

\subsection{Visualization of learned embeddings.}
\begin{figure*}[h!]
    \centering
    \includegraphics[width=0.95\textwidth]{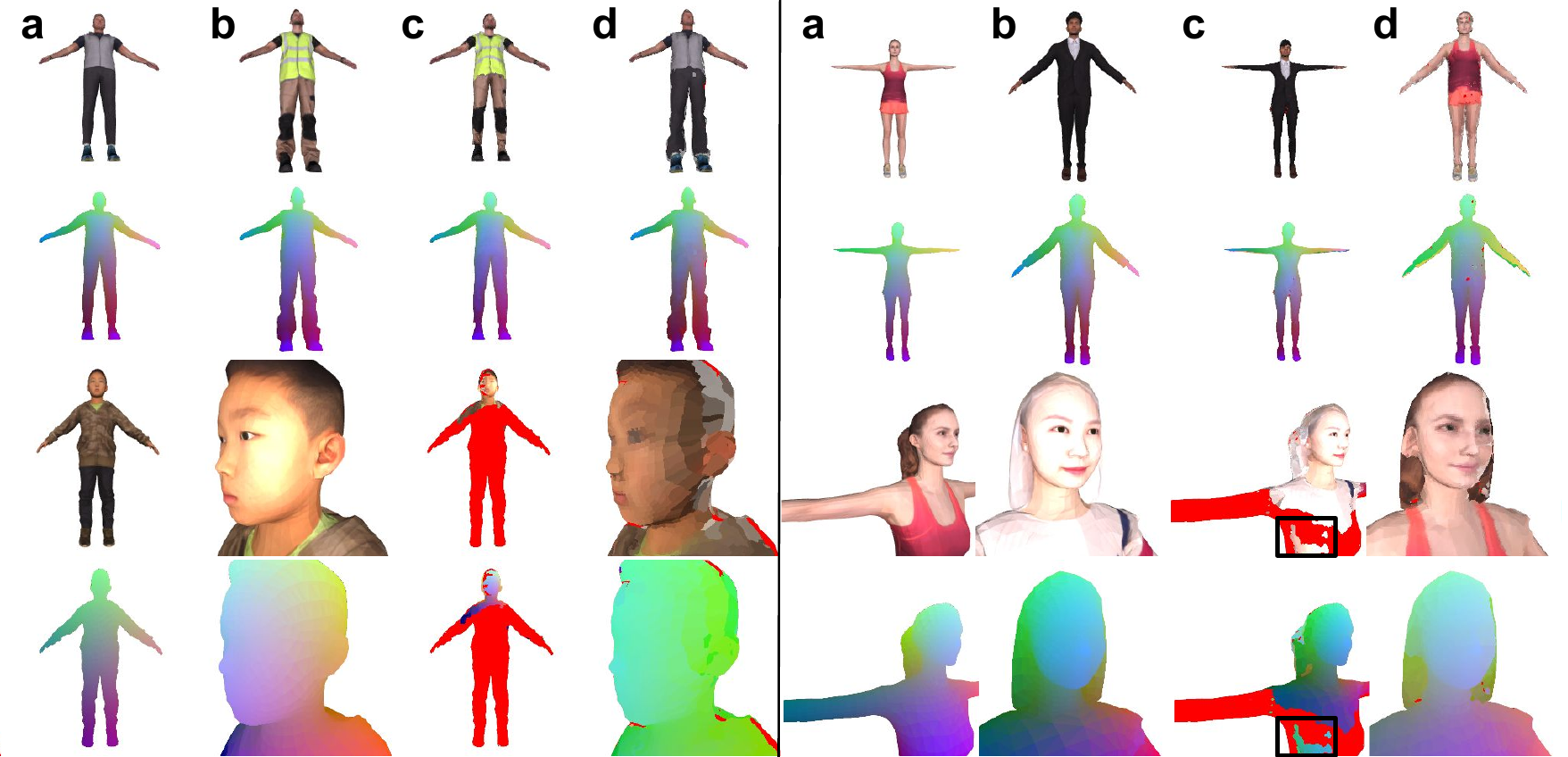}
    \caption{\textbf{Visualization of learned embeddings.} Given a
      pair of images in (a) and (b), our network produces per-pixel
      embedding for each image. 
      We map pixels from (b) to (a) according to feature similarity, resulting in (c).
      Similarly (d) is generated by
      transferring texture from (a) to (b). For
      pixels which have similarity below a threshold are colored
      red. We visualize the smoothness of our learned
      correspondence in the second and forth row. 
      Our method learns to produce correct correspondences between human subject in different clothing and same
human subject in different camera poses (left). Our approach also finds correct correspondences between
different human subjects in different poses (right). Mistakes are highlighted using black boxes.  }
    \label{fig:correspondence}
\end{figure*}
Our self-supervision is based on enforcing consistency in pixel
embeddings across views for pixels that corresponds to the same point
in 3D. In Figure~\ref{fig:correspondence} we visualize correspondences 
using our learned embeddings between human subjects from the RenderPeople 
dataset in different costumes and poses. The smoothness and consistency 
in correspondences implies that the network can be fine-tuned with few
labeled examples and still perform well on unseen examples.
\begin{table*}[h]
\resizebox{\textwidth}{!}{%
\begin{tabular}{l|c|c|c|c|c|c|c|c|c|c|c|c|c|c|c}
\textbf{Methods} & \textbf{Mean} & \textbf{Fau.} & \textbf{Vase} & \textbf{Earph.} & \textbf{Knife} & \textbf{Bed} & \textbf{Bot.} & \textbf{Dishw.} & \textbf{Clock} & \textbf{Door} & \textbf{Micro.} & \textbf{Fridge} & \textbf{Stor.F.} & \textbf{Trash}  & \textbf{Dis}\\
\# semantic parts &  & 12 & 6 & 10 & 10 & 15 & 9 & 7 & 11 & 5 & 6 & 7 & 24 & 11 & 4\\
\hline
(2D) Scratch        & 32.0 & 29.9 & 31.5 & 32.4 & 25.0 & 27.3 & 30.5 & 34.1 & 19.2 & 26.7 & 35.1 & 26.8 & 19.3 & 33.6 & 76.4\\
(2D) ImageNet        & 32.8 & 30.1 & 34.5 & 33.1 & 23.8 & 29.2 & 30.8 & 32.9 & 20.1 & 28.1 & 36.1 & 27.5 & 19.6 & 34.7 & \textbf{78.7} \\
(2D) DenseCL~\cite{wang2020DenseCL}        & 34.2 &	\textbf{31.4} &	35.4&	33.6&	22.7&	30.8&	33.7&	36.7&	19.7&	28.9&	41.9&	30.2&	21.2&	34.7 & 78.3\\
\hline
(3D) Scratch       & 30.3 & 27.7 & 28.8 & 28.4 & 19.8 & 24.5 & 25.8 & 39.4 & 15.9 & 24.3 & 37.7 & 30.9 & 23.5 & 30.0 & 67.8\\
(3D) Deformation \cite{wang2020few} & 27.5 &	28.4&	27.2&	24.7&	20.6&	12.4&	34.7&	30.9&	17.4&	26.1&	38.8&	24.6&	14.2&	21.0&	63.8 \\
(3D) PointContrast~\cite{PointContrast2020} & 34.1 & 29.0 & 35.8 & 31.0 & \textbf{25.6} & 27.8 & 32.5 & 39.9 & \textbf{22.8} & \textbf{29.1} & 41.3 & \textbf{32.5} & \textbf{25.2} & 31.1 & 73.4\\
\hline
(2D+3D) \mvd          & \textbf{35.9} &	31.1 &	\textbf{39.1} &	\textbf{34.8} &	25.2 &	\textbf{32.4} &	\textbf{39.2} &	\textbf{40.0} &	20.7 &	28.7 &	\textbf{44.3} &	29.8 &	22.6 &	\textbf{36.3} & 78.2
\end{tabular}%
}
\caption{\textbf{Few-shot segmentation on the Partnet dataset with limited labeled shapes.} 10 fully labeled shapes are provided for training. Evaluation is done on the test set of PartNet using the mean part-iou metric (\%). Training is done per category separately. Results are reported by averaging over 5 random runs.}\label{table:partnetfewshape}
\end{table*}

\begin{table*}[h]
\resizebox{\textwidth}{!}{%
\begin{tabular}{l|c|c|c|c|c|c|c|c|c|c|c|c|c|c|c}
\textbf{Methods} & \textbf{Mean} & \textbf{Fau.} & \textbf{Vase} & \textbf{Earph.} & \textbf{Knife} & \textbf{Bed} & \textbf{Bot.} & \textbf{Dishw.} & \textbf{Clock} & \textbf{Door} & \textbf{Micro.} & \textbf{Fridge} & \textbf{Stor.F.} & \textbf{Trash}  & \textbf{Dis}\\
\# semantic parts &  & 12 & 6 & 10 & 10 & 15 & 9 & 7 & 11 & 5 & 6 & 7 & 24 & 11 & 4\\
\hline
(2D) Scratch        & 25.9 & 21.7 & 25.2 & 26.1 & 19.3 & 19.8 & 25.0 & 27.3 & 16.6 & 25.0 & 27.9 & 22.3 & 13.2 & 24.2 & 68.7\\
(2D) ImageNet      & 27.1 & 23.2 & 27.9 & 28.1 & 20.1 & 21.2 & 25.4 & 28.2 & 16.6 & 25.6 & 29.3 & 23.4 & 12.6 & 27.4 & 70.1\\
(2D) DenseCL~\cite{wang2020DenseCL}   & 28.9 &	23.9&	31.3&	29.0&	21.3&	22.4&	28.9&	29.6&	16.4&	27.4&	33.6&	25.0&	15.9&	28.4 & 71.9\\
\hline
(3D) Scratch        & 17.1 & 14.8 & 17.6 & 16.4 & 12.1 & 8.2  & 15.7 & 19.0 & 7.7  & 20.7 & 20.9 & 15.8 & 6.8  & 10.6 & 52.9\\
(3D) PointContrast~\cite{PointContrast2020} & 28.4 & 22.3 & 32.5 & 28.6 & 21.2 & 18.9 & 25.9 & \textbf{31.3} & \textbf{18.9} & \textbf{28.5} & 31.4 & 24.8 & 15.5 & 25.8 & \textbf{72.1}\\
\hline
(2D+3D) \mvd           & \textbf{30.3} & \textbf{25.5} & \textbf{33.7} & \textbf{31.6} & \textbf{22.4} & \textbf{24.9} & \textbf{31.7} & 31.0 & 16.2 & 25.8 & \textbf{35.7} & \textbf{25.6} & \textbf{17.0} & \textbf{31.4} & 71.2
\end{tabular}%
}
\caption{\textbf{Few-shot segmentation on the PartNet dataset with limited labeled 2D views.} 10 shapes, each containing $v=5$ random labeled views, are used for training. Evaluation is done on the test set of PartNet using the mean part-iou metric (\%). Training is done per category separately. Results are averaged over 5 random runs.}\label{table:partnetfewviews}
\end{table*}

\begin{table*}[h]
\centering
\resizebox{0.7\textwidth}{!}{%
\begin{tabular}{lc|c|c|c|c|c|c|c}
    \textbf{Methods}          &      \multicolumn{4}{c|}{\textbf{k=30, v=all}} & \multicolumn{4}{c}{\textbf{k=30, v=5}} \\
\hline
       & \textbf{Mean} & \textbf{Chair}     & \textbf{Table}    & \textbf{Lamp}    & \textbf{Mean}  & \textbf{Chair} & \textbf{Table} & \textbf{Lamp} \\
\# semantic labels &     &  39 &  51 & 41 &  & 39   &  51   & 41\\
\hline
(2D) Scratch    & 13.7 & 20.8      & 10.1     & 10.3    &    10.6   & 15.2  & 7.3   & 9.2  \\
(2D) Imagenet      & 13.7 & 20.4      & 9.9      & 10.8    & 11.4       & 16.4  & 7.7   & 10.1 \\
(2D) DenseCL \cite{wang2020DenseCL}       & 15.7 & 22.6      & 11.7     & \textbf{12.6}    & 12.0       & 17.1  & 8.5   & \textbf{10.4} \\
\hline
(3D) Scratch 3d    & 11.5 & 17.8      & 8.0      & 8.7     & 6.3       & 10.1  & 4.5   & 4.3  \\
(3D) Deformation \cite{wang2020few} & 6.5 & 8.4 & 4.9 & 6.1 & - & - & - & - \\
(3D) Pointcontrast \cite{PointContrast2020} & 14.5 & 23.0      & 10.8     & 9.8     & 11.8       & \textbf{20.4}  & 7.8   & 7.1  \\
\hline
(2D+3D) Ours          & \textbf{16.6} & \textbf{25.3}      & \textbf{12.9}     & 11.7    &   \textbf{12.8}    &   19.3    &   \textbf{9.8}    &    9.3 
\end{tabular}
}
\caption{\textbf{Few-shot segmentation on the PartNet dataset.} \textbf{Left:} $30$ fully labeled shapes are used for training. \textbf{Right:} $30$ shapes are used for training, each containing $v=5$ random labeled views. Evaluation is done on the test set of PartNet with the mean part-iou
metric (\%). Results are reported by
averaging over 5 random runs.
}\label{table:partnetfewshot30}
\end{table*}

\subsection{Few-shot segmentation on PartNet}\label{sec:partnet}
\paragraph{\textbf{\textit{k} fully labeled shapes.}}
Table~\ref{table:partnetfewshape} shows results using the part mIOU metric
on few-shot semantic segmentation on PartNet dataset using $k=10$ fully
labeled shapes. Our method performs better than the network trained
from scratch by $4\%$ (part mIOU) showing the effectiveness of our self-supervision approach. 
Our architecture initialized with ImageNet
pre-trained weights, improves performance over training from scratch,
implying pre-training on large labeled datasets is helpful even when
the domain is different. The DenseCL baseline, which is trained on our
dataset, improves performance over ImageNet pre-trained weights, owing
to the effectiveness of contrastive learning at instance level and at
dense level. Interestingly, 2D training from scratch performs better
than 3D training from scratch. The learned deformation based alignment approach \cite{wang2020few} performs worse because aligning shapes of different topology and structure does not align semantic parts well. Furthermore, alignment is agnostic to the difference in the set of fine-grained semantic parts between shapes. The 3D sparse convolution network
pre-trained using point contrastive learning on our dataset and
fine-tuned with few labeled shapes performs better than all previous
baselines. Finally our approach, that uses dense contrastive learning
at pixel level outperforms all baselines.
\paragraph{\textbf{Sparse labeled 2D views.}}
Table \ref{table:partnetfewviews} shows the results on few-shot
semantic segmentation on PartNet dataset using sparse 2D views
for supervision. Here, the DenseCL baseline outperforms training
from scratch and the ImageNet initialized network. DenseCL also
outperforms the 3D PointContrast baseline, showing the effectiveness of 2D
architectures and 2D self-supervision. Finally, our approach outperforms all baselines.
Note that evaluation on Chairs, Lamps and Tables categories is shown separately in Table \ref{table:partnetfewshot30} with $k=30$, because our randomly selected $k=10$ shapes do not cover all the part labels of these classes. In this setting our approach outperforms the baselines.
\subsection{Few-shot segmentation on RenderPeople}\label{sec:renderpeople}
\begin{figure*}[]
\centering
\includegraphics[width=0.95\textwidth]{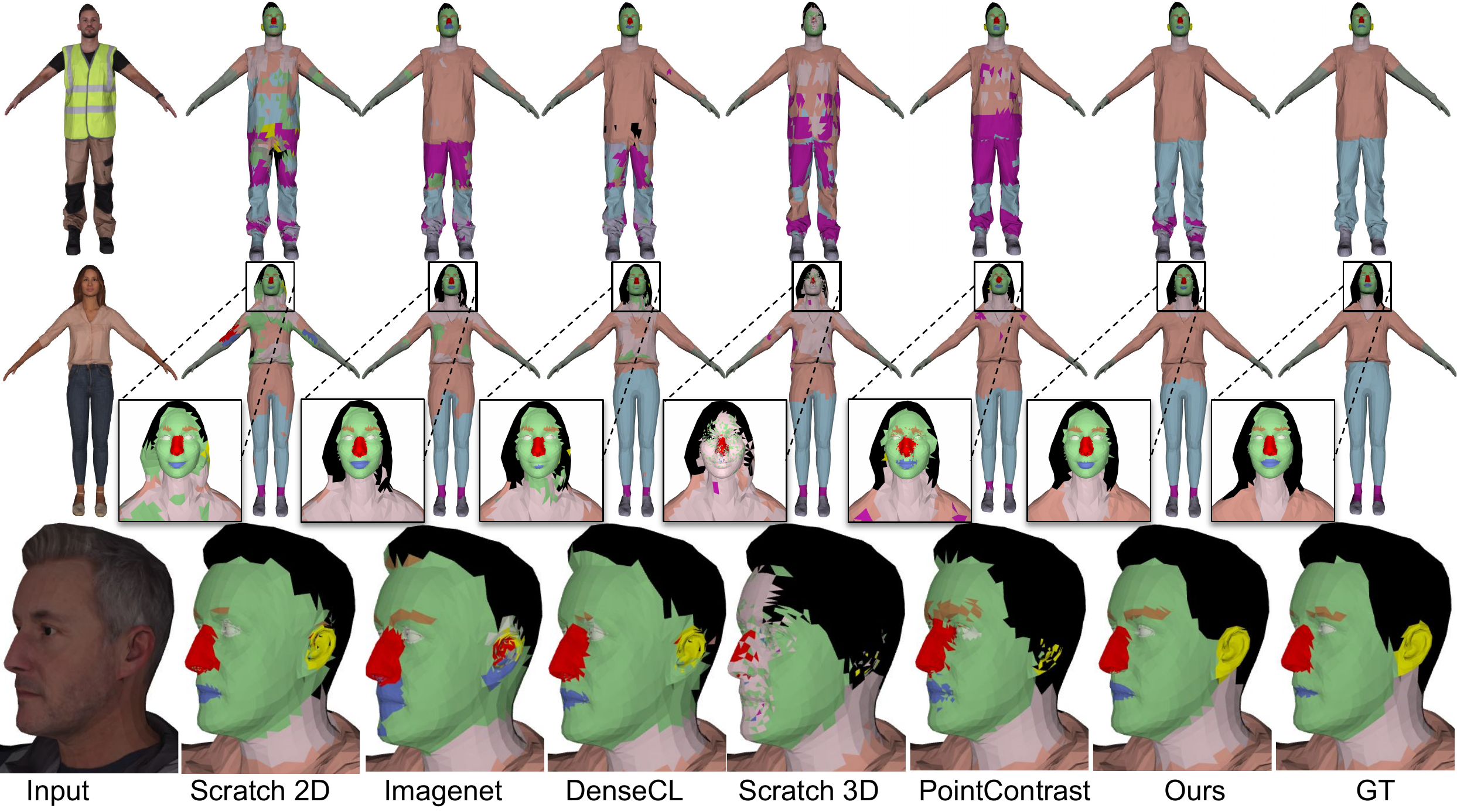}
\caption{\textbf{Visualization of predicted semantic labels on the Renderpeople dataset in the few-shot setting when $k=5$ fully labeled shapes are used for fine-tuning.} We visualize the predictions of all baselines. Our method produces accurate semantic labels for 3D shapes even for small parts, such as ears and eyebrows.}
\label{fig:renderpeopleresults}
\end{figure*}

To evaluate our method on the textured dataset, we use the RenderPeople
dataset. We use the same set of 2D and 3D baselines as described in \S \ref{sec:baselines}.
Note that, we provide color and normal with point cloud input to 3D Scratch and 3D PointContrast.
In addition to the settings described in \S \ref{sec:expsetting},
we analyze the effect of different inputs
given to the network, \ie when only RGB images are used as input for
self-supervision and fine-tuning, and when both RGB images and geometry
information (normal + depth maps) are available for self-supervision
and fine-tuning. We train all baselines in these two settings, except
3D baselines that take geometry by construction. 
The results are shown in  Table \ref{table:renderpeople}.
\begin{table}[t]
\centering
\resizebox{\textwidth}{!}{%
\begin{tabular}{c|c|c|c|c|c|c|c|c}
                                            \textbf{Methods} & \multicolumn{4}{c|}{\textbf{RGB}}   & \multicolumn{4}{c}{\textbf{RGB+Geom.}} \\
             & $k$=5, $v$=all & $k$=10, $v$=all & $k$=5,$v$=3   & $k$=10,$v$=3  & $k$=5, $v$=all & $k$=10, $v$=all & $k$=5,$v$=3   & $k$=10,$v$=3  \\
\hline
(2D) Scratch                                & 50.2 & 60.5 & 38.7 & 46.3 & 55.3  & 62.6  & 40.6  & 50.4  \\
(2D) ImageNet                               & 58.8 & 67.6 & 48.9 & 58.1 & 55.3  & 63.7  & 44.3  & 51.9  \\
(2D) DenseCL \cite{wang2020DenseCL}         & 58.3 & 66.8 & 46.5 & 55.5 & 56.0  & 64.0  & 31.0  & 41.5  \\
\hline
(3D) Scratch                                & -    & 48.1 & -    & 26.1 & 35.0  & -     & 14.5  & -     \\
(3D) PointContrast \cite{PointContrast2020} & -    & 61.3 & -    & 56.7 & 53.0  & -     & 48.5  & -     \\
\hline
(2D+3D) \mvd & \textbf{67.5}  & \textbf{73.8}   & \textbf{59.6} & \textbf{67.4} & \textbf{58.8}  & \textbf{65.0}   & \textbf{50.3} & \textbf{55.1}
\end{tabular}%
}
\caption{\textbf{Few-shot segmentation on the RenderPeople dataset.} We
evaluate the segmentation performance using the part mIOU metric. We experiment with two kinds of input, 1) when both RGB+Geom. (depth and normal maps) are input, and 2) when only RGB is input to the network. We evaluate all methods when $k=5,10$ fully labeled shapes are used for supervision and when $k=5,10$ shapes with 3 2D views are available for supervision. \mvd consistently outperform baselines on all settings.}
\label{table:renderpeople}
\end{table}
\paragraph{\textbf{RGB+Geom.}}
In the first setting when RGB is used as input along with geometry information (normal + depth), our
approach outperforms all the baselines, with $3.5\%$ and $9.7\%$ improvement on training from scratch when only $k=5$ labeled shapes are given and when $k=5$ shapes with $v=3$ views are given for supervision respectively. We use only $3$ views for RenderPeople dataset because of its simpler topology in comparison to $5$ views for PartNet. The ImageNet pre-trained model,
which is modified to take depth and normal maps as input performs similar
to training from scratch, that implies that the domain shift is too large
between ImageNet and our dataset. DenseCL applies dense correspondence
learning at a coarse grid and hence does not perform well in the dense
prediction task when only a few labeled examples are given.

\paragraph{\textbf{RGB only.}}
In the second setting, when only RGB image is input to the network, \mvd gives $17.3\%$ and $20.9\%$ improvement over training from scratch when only $k=5$ labeled shapes are given, and when $k=5$ shapes with $v=3$ views are given for supervision respectively. The ImageNet and DenseCL both perform better than training from scratch, including their counterpart which takes both RGB+geometry as input. 
\mvd with only RGB as input also performs significantly better than its RGB+geometry counterpart. We expect this
behaviour is due to the following reasons: first when geometry is also used as input
to the network, the pre-training task focuses more on geometry to
produce consistent embeddings, as is shown in Figure
\ref{fig:correspondence}, where consistent embeddings are produced
between the same human subject in two different costumes. However,
when only RGB is input to the network, the pretraining task focuses on
RGB color only to learn correspondences. Second, the semantic segmentation
of human models requires high reliance on RGB features compared to
 geometry, and the additional geometry input tends to confuse the pre-trained network.

\begin{figure}
    \centering
    \includegraphics[width=\linewidth]{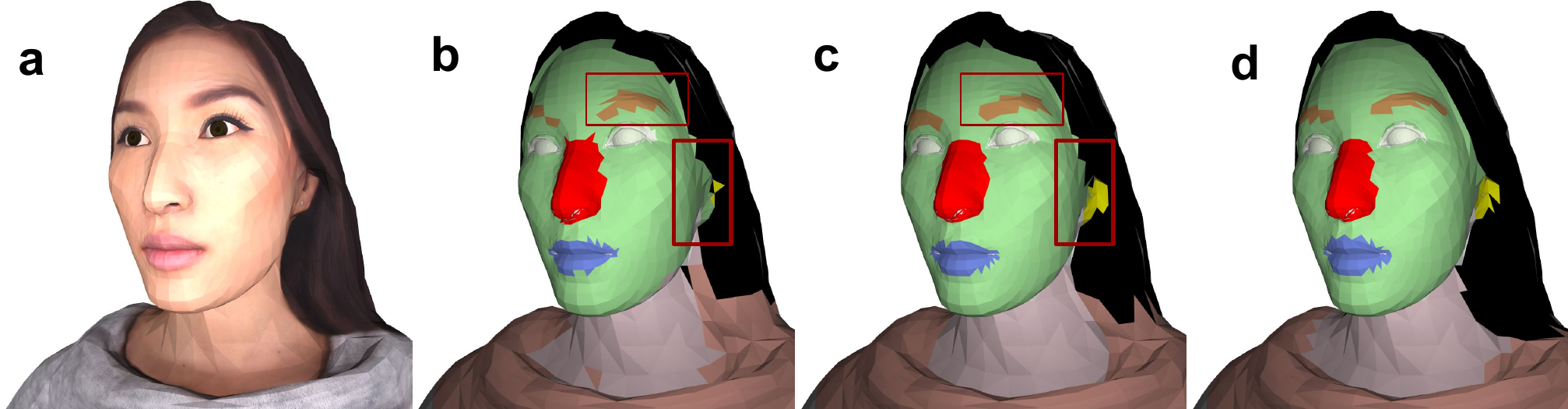}
    \caption{\textbf{View aggregation.} Given the input in (a), \mvd produces 2D labels (b) which can further be improved by multi-view aggregation (c) as is highlighted in boxes and produces segmentation close to the ground truth (d).}\label{fig:aggregation}
\end{figure}

Figure \ref{fig:renderpeopleresults} shows qualitative results of different methods. \mvd consistently outperforms all baselines and can segment tiny parts such as eyes, ears and nose. We also refer to the Supplement for more qualitative visualization. Figure \ref{fig:aggregation} shows the effect of multi-view aggregation on 3D segmentation.


\begin{table}[t]
\centering
\resizebox{0.5\textwidth}{!}{%
\begin{tabular}{l|c|c}
Method               & RGB+Geom. & RGB \\
\hline
\mvd w/o closeup views &  51.6 & 57.4 \\
\mvd w/o reg. & 58.0  & 67.2 \\
\mvd  &  58.8     & 67.5 \\    
\end{tabular}
}
\caption{\textbf{Effect of renderings and regularization on the RenderPeople dataset.} \mvd without closeup views for pre-training and fine-tuning performs worse compared to using closeup views. Our regularization term in the loss also shows improvement.}\label{table:viewselection}
\end{table}
\paragraph{\textbf{Regularization.}}
During the fine-tuning stage, we use an extra regularization term $\lambda {\cal L}_{\textsf{ssl}}$ applied on shapes from $\mathcal{X}_u$, to prevent the network from overfitting on the small training set $\mathcal{X}_l$ as described in \S \ref{sec:segmentation}. In Table \ref{table:viewselection}, we show that this regularization improves our performance on the RenderPeople dataset.

\paragraph{\textbf{Effect of view selection.}}
We also analyze the effect of view selection. In our previous experiment, we select views by placing camera farther away from the shape to obtain a full context along with placing the camera close to the shape to obtain finer details. The case of removing close-up views during pre-training and fine-tuning stage is examined in Table \ref{table:viewselection}. We observe that closeup views are important for accurate segmentation of small parts. Finally, we also observe that on the RenderPeople dataset, the segmentation performance improves as more views are provided during inference. As the number of views are increased from $5$ to $96$, the segmentation performance improves from $54.7\%$ to $58.8\%$.

%% file: conclusion.tex
\section{Conclusion}
In this paper, we present \mvd, a self-supervision method that learns a multi-view representation for 3D shapes with geometric consistency enforced across different views.  We pre-train our network with a multi-view dense correspondence learning task, and show that the learned representation outperforms state-of-art methods in our experiments of few-shot fine-grained part segmentation, giving most benefits for textured 3D shapes.
\paragraph{\textbf{Limitations.}} Our method relies on 2D renderings of 3D shapes, thus a few surface regions may not be covered due to self-occlusion. In this case, the label predictions in these regions are unreliable.
Our self-supervision also requires rendering several views of shapes to make the representations view invariant, which increases the computational cost. Our view selection during pre-training and fine-tuning stage is heuristic-based and can be improved by a learnable approach \cite{hamdi2021mvtn}. A useful future avenue is to combine our approach of 2D correspondence learning with 3D correspondence learning \cite{PointContrast2020} to obtain the best of both worlds.
\mvd may also open up other potential supervision sources, such as reusing existing image segmentation datasets to segment 3D shapes, exploiting motion in videos to provide correspondence supervision for pre-training.

\paragraph{Acknowledgements.} Subhransu Maji acknowledges support from NSF grants \#1749833
and \#1908669.

%% file: supp.tex
\appendix
\section{Supplementary Material}
Here we further provide the following supplementary information and results:
\begin{itemize}
    \item Samples from training datasets
    \item Experiment on the Shapenet dataset
    \item Additional qualitative results on RenderPeople, and details of annotation for the dataset
    \item Training details of our approach and baselines
    \item Data licenses, and discussion on societal impact and human dataset.
\end{itemize}

\subsection{Samples from training datasets}
\begin{figure}
    \centering
    \includegraphics[scale=0.3]{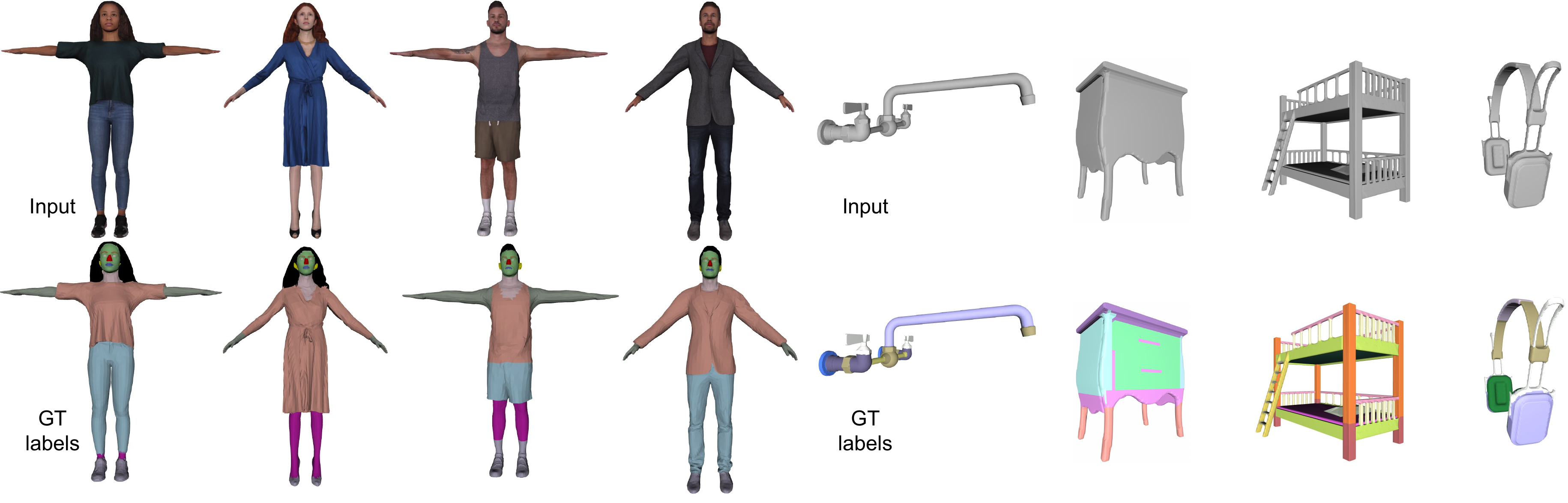}
    \caption{\textbf{Examples from datasets used in our experiments.} \emph{Left}: RenderPeople~\cite{Renderpeople} dataset. \emph{Right}: PartNet~\cite{partnet}  dataset\label{fig:datasets} }
\end{figure}
Figure \ref{fig:datasets} visualizes different samples from our training dataset used in our experiments.

\subsection{Experiment on Shapenet dataset}
The main focus of our work is fine-grained semantic segmentation. We also experiment with the Shapenet Semantic Segmentation dataset \cite{Chang2015ShapeNetAI} for the task of few-shot semantic segmentation, which consists of $16,881$ labeled point clouds across 16 shape categories, with a total of $50$ part categories. We transfer the point labels to triangles of a mesh using nearest neighbor queries to train our models. The evaluation is done by transferring the predicted triangle labels back to original point cloud. We use the same architecture and training strategy for this dataset as used for other datasets. We report our results in Table \ref{table:sota_shapenet}. Note that instance
mIOU is highly influenced by the shape categories with large number of
testing shapes,  e.g., Chair, Table.  Class mIOU, on the other hand gives
equal importance to all categories, hence it is a more robust
evaluation metric. We evaluate the performance of work by Wang \etal \cite{wang2020few} all all shape categories from this dataset and average the performance over 5 random runs.

\begin{table}[t]
\centering
\begin{tabular}{lc|c}
Methods    & instance avg. mIOU & class avg. mIOU\\
\hline
 SO-Net~\cite{li2018so}                      &  64.0 & -\\
 PointCapsNet~\cite{zhao20193d}              &  67.0 & -\\
 MortonNet~\cite{MortonNet}                  &  -    & -\\
 JointSSL~\cite{jointssl}                    &  71.9 & -\\
 Multi-task~\cite{hassani2019unsupervised}   &  68.2 & -\\
 Deformation \cite{wang2020few}              & 68.9              & 66.2 \\
PointContrast ~\cite{PointContrast2020}      &  74.0 & - \\
 ACD \cite{selfsupacd}                       &  \textbf{75.7} & 74.1\\ 
\hline
2D Scratch                                   & 72.7& 74.7  \\
(2D+3D) Ours                                         & 74.3& \textbf{75.8}\\
\end{tabular}
\caption{\small{\textbf{Comparison with state-of-the-art few-shot part segmentation methods on ShapeNet.} Performance is evaluated using \textit{instance-averaged} and \textit{class-averaged} mIOU while using $1\%$ of the training data.
}}\label{table:sota_shapenet}
\end{table}

\subsection{Additional results on RenderPeople and annotation details}

\begin{figure*}
    \centering
    \includegraphics[width=\textwidth]{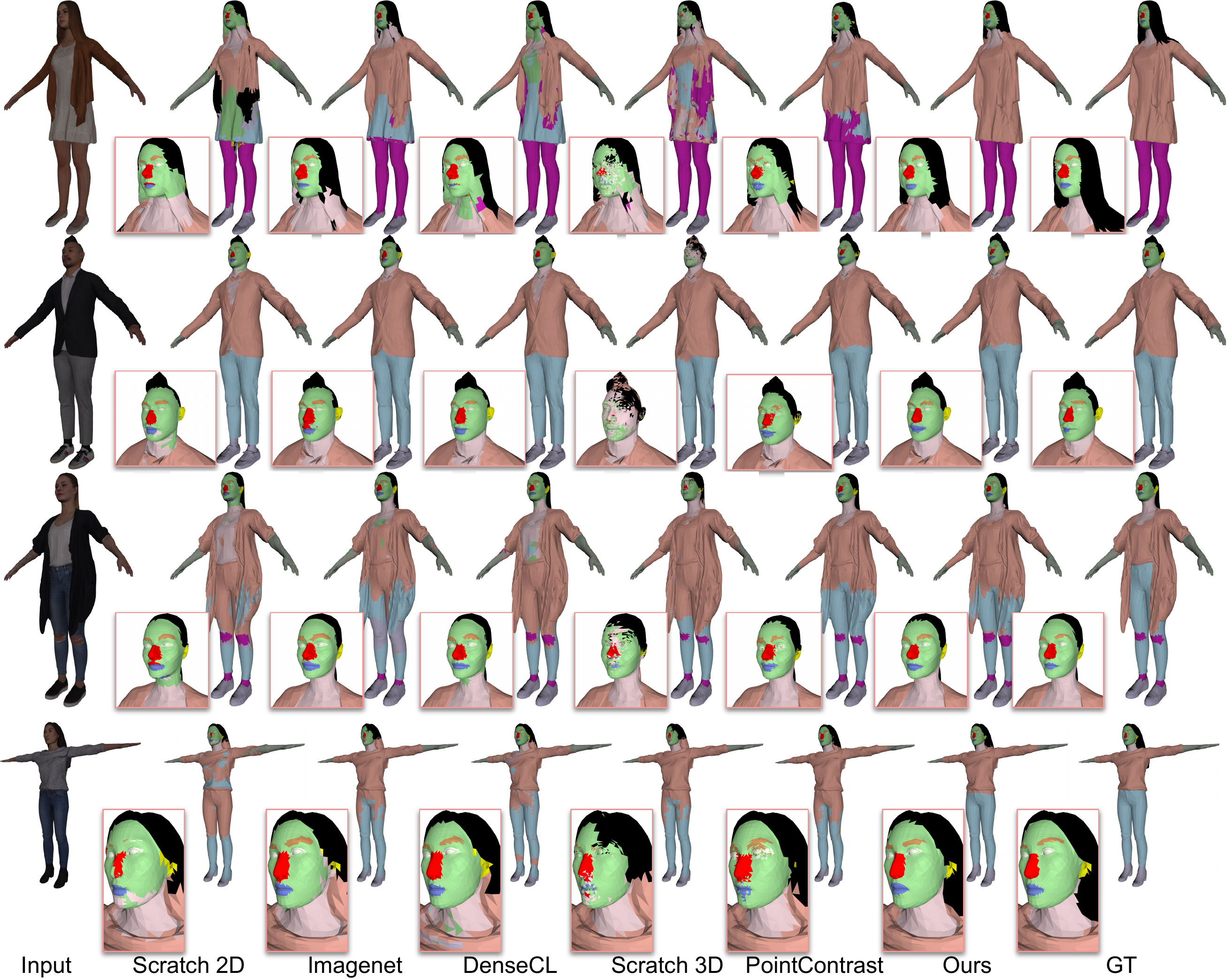}
\caption{\textbf{Visualization of predicted semantic labels on Renderpeople dataset in the few-shot setting when $k=5$ fully labeled shapes are used for fine-tuning.} We visualize the predictions of all baselines. To visualize the details of predicted segmentations in the facial region, we provide an inset figure.}
    \label{fig:segmentrenderpeople}
\end{figure*}

In Figure \ref{fig:segmentrenderpeople}, we provide additional qualitative results on the RenderPeople dataset.

\begin{figure}
    \centering
    \includegraphics[width=\linewidth]{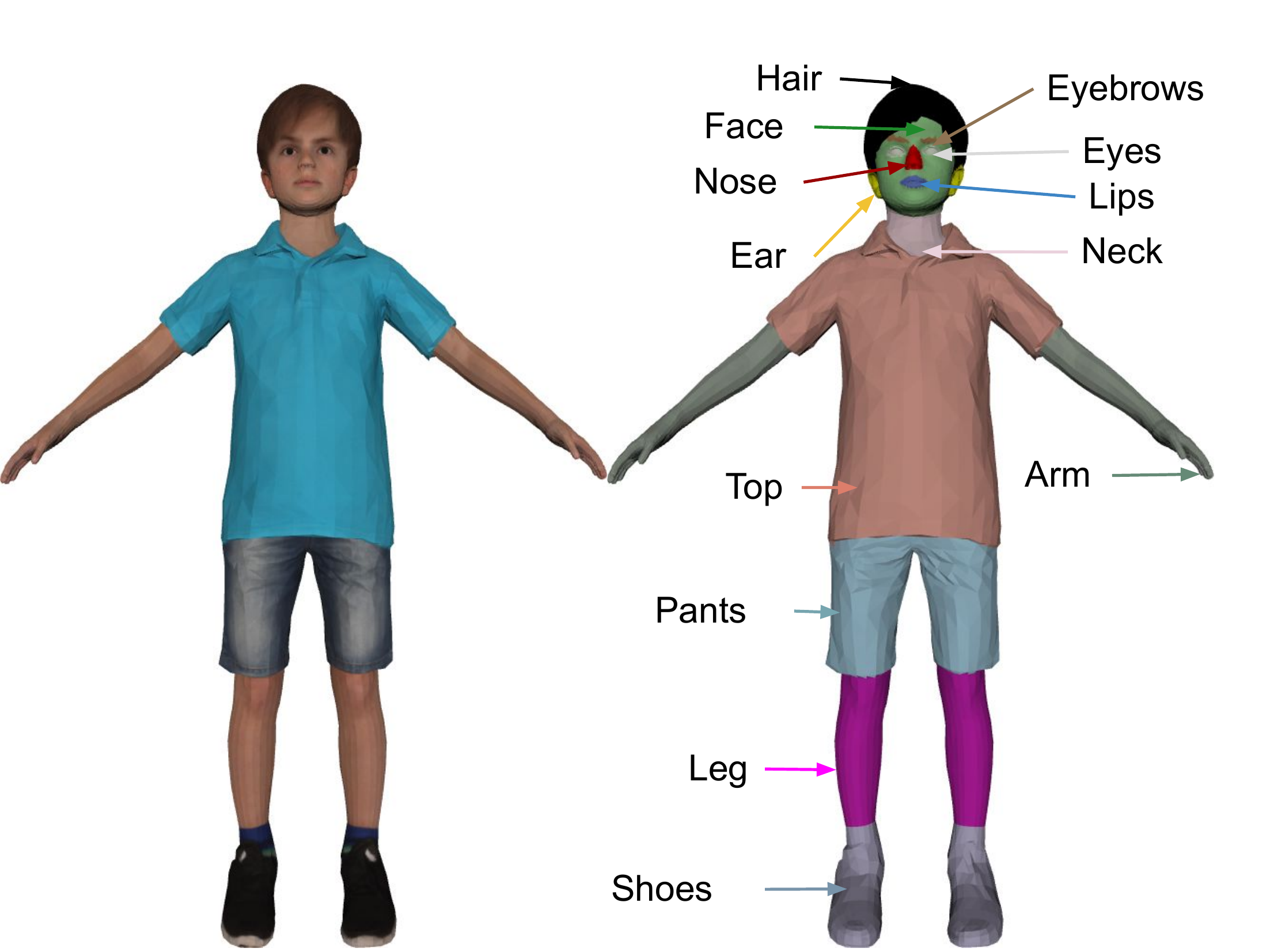}
    \caption{Semantic labels of a shape from the RenderPeople dataset.}
    \label{fig:labelrenderpeople}
\end{figure}

In the data annotation stage, we label Renderpeople shapes using the labeling tool from \cite{labelme2016}. We start by rendering multiple RGB images of the textured mesh such that maximum surface area can be covered. Then we label each rendered image and back-project the pixel labels to the surface. We label 13 different parts as shown in Figure \ref{fig:labelrenderpeople}.



\subsection{Training details}
\paragraph{Training details for PartNet dataset.}
For pre-training and fine-tuning stages of our method we use the Adam optimizer with $0.001$ learning rate. For pre-training we decay the learning rate by half when validation loss saturates. During pre-training, we use $4k$ matched pairs of points for a pair of views to compute our self supervision loss. During pre-training on the PartNet dataset, we train our model with batch size of $16$ for $200k$ iterations. For fine-tuning, we use the batch size of $8$ and exponential learning rate decay (factor=$0.99$) after every $40$ iterations. For $k=10$, $v=all$ setting, we train our model for $4k$ iterations, and for $k=10$ and $v=5$ setting, we train our model for $2k$ iterations.

\paragraph{Training details for RenderPeople dataset.}
For pre-training and fine-tuning stages of our method, we use the Adam optimizer with $0.001$ learning rate. For pre-training, we decay the learning rate by half when validation loss saturates. During pre-training, we use $4k$ matched pairs of points for a pair of views to compute our self supervision loss. During pre-training for the RenderPeople dataset, we train our model with batch size of $16$ for $100k$ iterations. For fine-tuning, we use the batch size of $8$ and exponential learning rate decay (factor=$0.99$) after every $40$ iterations. For the RenderPeople dataset for $k=5$, $v=all$ setting we train our model for $2K$ iterations, and for $k=5$ and $v=3$ setting, we train our model for $400$ iterations. 

\paragraph{DeepLabv3+.}
We use the DeepLabV3+ as our 2D CNN backbone for learning pixel level features. We modify the last layer of DeepLabV3+. In the original version, the $(64\times 64)$ feature map is directly $4\times$ upsampled to a $(256\times 256$) feature map using bilinear interpolation, since the input image has a size of $(256\times 256\times 3)$. We instead gradually upsample the $(64\times 64)$ feature map to $(256\times 256)$ resolution in two upsampling stages to preserve fine-grained details in the following way: $\texttt{Up}(2) \rightarrow \texttt{BN}(256) \rightarrow \texttt{Relu} \rightarrow \texttt{Conv2D}(256, 128, 3) \rightarrow \texttt{Up}(2) \rightarrow \texttt{BN}(256) \rightarrow \texttt{Relu} \rightarrow \texttt{Conv2D}(128, 64, 3)$. We also use bilinear up-sampling. $\texttt{Conv2D}(i,o,k)$ is a 2D convolution layer with $i$ input channels, $o$ output channels and $k$ kernel size, $\texttt{Relu}$ is rectified linear unit, $\texttt{Up}(x)$ is bilinear up-sampling by a factor of $x$ and $\texttt{BN}$ is a batch normalization layer. 

\paragraph{DenseCL.}
We keep all the hyper parameters same as proposed in the original work. When depth map and normal maps are also input to the network, the spatial augmentations applied to the RGB image are also applied to the normal and depth maps. We do not augment normal and depth maps in any other way. The models are trained until convergence. Once the DenseCL baseline is pre-trained using their proposed approach on our dataset, we use the backbone ResNet weights to initialize our DeepLabv3+ architecture as described above and add a  2D convolution layer as a segmentation head. 

\paragraph{PointContrast.}
To implement our 3D baseline, we use a 3D ResNet with U-Net based architecture with 42 layers as proposed in the original paper \cite{PointContrast2020}. We use a voxel size of $0.01$. We use a batch size of $16$ and $10k$ pairs of matched points to compute their self supervision loss. The implementation of the loss is done using the source code provided by the authors. We use the SGD optimizer with learning rate $0.1$ with $0.9$ momentum and $0.0001$ weight decay. We train this model for $100k$ iterations. The validation loss saturates after $100k$ iterations.

\subsection{Dataset, code, and ethics discussions}

\subsubsection{Dataset and code licenses}
The PartNet dataset~\cite{partnet} is a collection of labeled shapes from ShapeNet~\cite{Chang2015ShapeNetAI}. The license can be found on the website of ShapeNet.
We obtained the license for using the Renderpeople dataset~\cite{Renderpeople} through an agreement with Renderpeople. To run the comparison with baseline methods, we use the source code provided by the authors of DenseCL~\cite{DenseCL_git}, PointContrast~\cite{PointContrast_git}. The licenses of the codebases are provided in their original GitHub repositories.

\subsubsection{Potential negative societal impacts}
We present a method for labeling detailed parts of 3D models given a provided training set of shapes.  Like many other learning-based methods, our results can be biased by training datasets.  For purpose of deploying the method for human shapes,  one would need to carefully de-bias the dataset to cover the distribution of a wide range of body shapes, clothing,  skin tones, race, and gender. 

\subsubsection{Personal data and human subjects}
Our paper uses human 3D models from Renderpeople for training and evaluation.   The data collection and ethics approvals were taken care of by the dataset provider.  We carefully inspected the dataset and did not find identifiable information or offensive content. More information about the dataset can be found on the websites of the data provider.

